\documentclass[journal]{IEEEtran}
\usepackage{ amssymb   }
\usepackage{ mathdots  }
\usepackage{ amsmath   }
\usepackage{ color     }
\usepackage{ graphicx  }
\usepackage{ amsfonts  }
\usepackage{ amssymb   }
\usepackage{ epstopdf  }
\usepackage{ upgreek   }
\usepackage{ bm        }
\usepackage{ graphicx  }
\usepackage{ subfigure }
\usepackage{ mathtools }
\usepackage{ algorithm }
\usepackage{algorithmic}
\usepackage{ booktabs  }
\usepackage{ multirow  }
\usepackage{ caption   }
\usepackage{ array     }
\usepackage{ float     }
\usepackage{   soul    }
\usepackage{ verbatim }
\usepackage[normalem]{ulem}

\usepackage[final]{pdfpages}
\usepackage[noadjust]{cite}
\newcolumntype{L}[1]{>{\raggedright\let\newline\\\arraybackslash\hspace{0pt}}m{#1}}
\newcolumntype{C}[1]{>{\centering\let\newline\\\arraybackslash\hspace{0pt}}m{#1}}
\newcolumntype{R}[1]{>{\raggedleft\let\newline\\\arraybackslash\hspace{0pt}}m{#1}}

\newtheorem{Lemma}{Lemma}

\newtheorem{Theorem}{Theorem}

\newtheorem{Assumption}{Assumption}

\newcommand{\xupdownarrow}[1]{%
	{\left\updownarrow\vbox to #1{}\right.\kern-\nulldelimiterspace}
}

\newcommand{\bv}{\boldsymbol{v}}
\newcommand{\bd}{\boldsymbol{d}}

\newcommand{\bxi}{\boldsymbol{\xi}}
\newcommand{\bnu}{\boldsymbol{\nu}}
\newcommand{\bmu}{\boldsymbol{\mu}}
\newcommand{\bbR}{\mathbb{R}}

\newcommand{\bbE}{\mathbb{E}}
\newcommand{\cS}{\mathcal{S}}
\newcommand{\cA}{\mathcal{A}}

\newcommand{\cP}{\mathcal{P}}

\newcommand{\cF}{\mathcal{F}}
\newcommand{\cV}{\mathcal{V}}
\newcommand{\cU}{\mathcal{U}}

\newcommand{\cO}{\mathcal{O}}
\newcommand{\suma}{\sum_{a \in \cA}} 
\newcommand{\sumi}{\sum_{i \in \cS}} 
\newcommand{\sumj}{\sum_{j \in \cS}}

\newcommand{\tmix}{t^*_{\text{mix}}}
\newcommand{\norm}[1]{\left\lVert#1\right\rVert}

\definecolor{antiquefuchsia}{rgb}{0.57, 0.36, 0.51}
\definecolor{cadmiumred}{rgb}{0.89, 0.0, 0.13}

\begin{document}
	\title{Voting-Based Multi-Agent Reinforcement Learning for Intelligent IoT}
	\author{\IEEEauthorblockN{
			Yue Xu$^{1,4}$, Zengde Deng$^{2}$, Mengdi Wang$^{3}$, Wenjun Xu$^{1}$, Anthony Man-Cho So$^{2}$, Shuguang Cui$^{4}$\bigskip}
		
		\IEEEauthorblockA{
			$^{1}$Key Lab of Universal Wireless Communications, Ministry of Education \\
			Beijing University of Posts and Telecommunications\\
			$^{2}$Department of Systems Engineering and Engineering Management, \\
			The Chinese University of Hong Kong, Hong Kong \\
			$^{3}$Department of Operations Research and Financial Engineering, Princeton University \\
			$^{4}$Shenzhen Research Institute of Big Data and The Chinese University of Hong Kong, Shenzhen\\}
	}
	
	\maketitle
	
	\begin{abstract}
		The recent success of single-agent reinforcement learning (RL) in Internet of things (IoT) systems motivates the study of multi-agent reinforcement learning (MARL), which is more challenging but more useful in large-scale IoT.
		In this paper, we consider a voting-based MARL problem, in which the agents vote to make group decisions and the goal is to maximize the globally averaged returns. 
		To this end, we formulate the MARL problem based on the linear programming form of the policy optimization problem and propose a primal-dual algorithm to obtain the optimal solution. 
		We also propose a voting mechanism through which the distributed learning achieves the same sublinear convergence rate as centralized learning. 
		In other words, the distributed decision making does not slow down the process of achieving global consensus on optimality. Lastly, we verify the convergence of our proposed algorithm with numerical simulations and conduct case studies in practical multi-agent IoT systems.
	\end{abstract}
	
	\section{Introduction}
	Reinforcement learning (RL) aims at maximizing a cumulative reward by selecting a sequence of optimal actions to interact with a stochastic \textit{unknown} environment, where the dynamics is usually modeled as a Markov decision process (MDP)~\cite{sutton2018reinforcement}. 
	Recently, single-agent RL has been successfully applied to contribute adaptive and autonomous intelligence in many Internet of things (IoT) applications, including smart cellular networks~\cite{8473693,8664581,xu2019load}, smart vehicle networks~\cite{8633948,8944302,8798668}, and smart unmanned aerial vehicles (UAV) networks~\cite{8432464,8377340,8672604}.
	Despite these successes, many recent studies envision that the IoT entities, e.g., smartphones, sensors, and UAVs, will become more decentralized, ad-hoc, and autonomous in nature~\cite{8714026,8166730}. 
	This encourages the extension from single-agent RL to multi-agent RL (MARL) to study the smart collaboration among local entities in order to deliver a superior collective intelligence, instead of simply treating them as independent learners.
	However, MARL is more challenging since each agent interacts with not only the environment but also the other agents. 
	
	
	Although a number of collaborative learning models based on MARL have been recently proposed~\cite{8629363,8792117,8807386,foerster2016learning, gupta2017cooperative, lowe2017multi, omidshafiei2017deep, foerster2017stabilising, zhang2018fully}, they usually impose a discount factor $ \gamma \in (0,1) $ on the future rewards to render the problem more tractable, e.g., bounding the cumulative reward~\cite{6949624, wai2018multi, lee2018primal}.
	However, many optimization tasks in the IoT systems, e.g., resource allocation and admission control, are long-run or non-terminating tasks. Existing studies reveal that the RL methods based on discounted MDP may yield a poor performance in the continuing tasks and become computationally challenging when the discount factor is close to one~\cite{sutton2018reinforcement,yang2017average,Yang2016Efficient,Ghavamzadeh2007Hierarchical}. This necessitates the development of MARL models based on the undiscounted average-reward MDP (AMDP) to tackle the continuing optimization tasks in IoT systems.
	Moreover, existing MARL models usually exhibit a performance degradation compared with their centralized versions~\cite{7500084,zhang2018fully} and only provide asymptotic convergence to an optimal point~\cite{7500084,zhang2018fully} or simply give empirical evaluations without theoretical guarantees~\cite{foerster2016learning, gupta2017cooperative, lowe2017multi, omidshafiei2017deep, foerster2017stabilising}.
	In contrast, in this paper, we give a sublinear convergence rate and theoretically prove that our proposed MARL model achieves the same convergence rate as centralized learning, which makes it a decent learning paradigm for distributed IoT systems.
	
	Meanwhile, it is critical to specify a proper collaboration protocol in order to promote safe and efficient cooperations in MARL systems.
	Many existing MARL models are built upon the centralized learning with decentralized execution framework where the agents perform iterative parameter consensus with a centralized server~\cite{lowe2017multi,8792117,Landon2016Multi,Nguyen2020Deep}. Moreover, the centralized server is assumed to have access to the behavioral policy or value functions of all distributed agents for model training. However, in many IoT applications~(e.g., location services), the privacy-sensitive data~(e.g., policy or value functions) should not be logged onto a centralized center due to privacy and security concerns. 
	On the other hand, recent works also propose a number of decentralized solutions which coordinate the agents through iterative parameter consensus among neighboring agents~\cite{zhang2018fully, wai2018multi, 6949624, lee2018primal}. However, this may give rise to massive communication overhead in large-scale IoT networks. Besides, their convergence depends on the connectivity properties of the networked agents, which can be topology prohibitive in a randomly deployed IoT network.
	The above issues motivate us to propose a new collaboration protocol for MARL which can coordinate the local entities in a safe and communication-efficient way.

	In this paper, we consider a collaborative MARL setting where the agents vote to make group decisions and the aim is to maximize the globally averaged return of all agents in the environment. Our primary interest is to develop a sample-efficient model-free MARL algorithm built upon voting-based coordinations in the context of infinite-horizon AMDP.
	Particularly, the considered AMDP does not assume the future rewards to be discounted while only needing to satisfy certain fast mixing property. This significantly complicates our analysis when compared with the discounted cases.
	The main contributions are summarized as follows. 
	\begin{itemize}
		\item We formulate the MARL problem in the context of AMDP based on the linear programming form of the policy optimization problem and propose a primal-dual algorithm to obtain the optimal solution.
		\item We provide the first sublinear convergence rate for solving the MARL problem for infinite-horizon AMDP. The proposed algorithm and theoretical analysis also cover the single-agent RL as a special case, which makes them more general.
		\item We propose a voting-based collaboration protocol for the proposed MARL algorithm, through which the distributed learning achieves the same sublinear convergence as centralized learning. In other words, the proposed distributed decision-making process does not slow down the process of achieving global optimality. Moreover, the proposed voting-based protocol has superior data privacy and communication-efficiency than existing parameter-consensus-based protocols.
	\end{itemize}
	In addition, we also verify the convergence of our proposed algorithm through numerical simulations and conduct a case study in a multi-agent IoT system to justify the learning effectiveness.
	
	The proposed model is promising for solving the long-run or non-terminating optimization tasks in multi-agent IoT systems, where distributed agents vote to determine a joint action, aiming at maximizing the globally averaged return of all agents. For example, the model can be employed to learn the optimal resource~(e.g., communication bandwidth and channel) allocation policy for a group of IoT devices to improve the overall capacity; learn the optimal on/off policy for a group of base stations to improve the overall energy efficiency; learn the optimal trajectory planning policy for a group of UAVs to avoid collisions.
	Moreover, since the distributed agents only need to exchange their vote information for collaboration, without revealing their policy or value functions to each other, the proposed model would be preferable in privacy-sensitive applications, e.g., location services.

	The remainder of this paper is organized as follows. 
	Section~\ref{sec:related_work} reviews the existing works on MARL.
	Section~\ref{sec:problem_formulation} introduces the problem formulations.
	Section~\ref{sec:algorithm} presents the voting-based multi-agent reinforcement learning algorithm.
	Section~\ref{sec:theoretical} presents the convergence analysis of our proposed algorithm.
	Section~\ref{sec:simulation} discusses the simulation results. 
	Finally, Section~\ref{sec:conclusion} concludes the paper.
	
	\textbf{Notation:} For a vector $\boldsymbol{x} \in \bbR^n$, we denote its $i$-th component as $x_i$, its transpose as $\boldsymbol{x}^{\top}$, and its Euclidean norm as $\|\boldsymbol{x}\|=\sqrt{\boldsymbol{x}^{\top}\boldsymbol{x}}$. 
	For a positive number $x$, we write $\log x$ for its natural logarithm.
	For a vector $ \boldsymbol{e} = (1, \ldots, 1)^{\top} $, we denote by $ \boldsymbol{e}_i $ the vector with its $ i $-th entry equaling $ 1 $ and other entries equaling $ 0 $.
	For two probability distributions $p,q$ over a finite set $X$, we denote their Kullback-Leibler (KL) divergence as $D_{KL}(p||q) = \sum_{x\in X} p(x) \log \frac{p(x)}{q(x)}$. 
	
	\section{Related Work}
	\label{sec:related_work}
	Many existing model-free MARL algorithms are based on the framework of Markov games~\cite{panait2005cooperative, Littman1994, littman2001friend, yang2018mean, 7534842} or temporal-difference RL~\cite{zhang2018fully, foerster2016learning, gupta2017cooperative, lowe2017multi, omidshafiei2017deep, foerster2017stabilising}.
	In the context of Markov games, the study of MARL usually models the MARL as stochastic games, such as cooperative games~\cite{panait2005cooperative}, zero-sum stochastic games~\cite{Littman1994,Vamvoudakis2017Game,Tamimi2007Model,Kim2010Model}, general-sum stochastic games~\cite{littman2001friend}, decentralized Q-Learning~\cite{7534842}, and the recent mean-field MARL~\cite{yang2018mean}.
	Alternatively, the study of MARL in the context of temporal-difference RL mainly originates from dynamic programming, which learns by following the Bellman equation, including the ones based on deep neural networks~\cite{foerster2016learning, gupta2017cooperative, lowe2017multi, omidshafiei2017deep, foerster2017stabilising} and the ones based on linear function approximators~\cite{zhang2018fully}.
	However, first, the above MARL models can only provide asymptotic convergence~\cite{zhang2018fully} to an optimal point or simply provide empirical evaluations without theoretical guarantees~\cite{foerster2016learning, gupta2017cooperative, lowe2017multi, omidshafiei2017deep, foerster2017stabilising}. Second, they are all based on the discounted MDP, instead of the undiscounted AMDP.
	On the other hand, though average-reward RL has received much attention in recent years, most of them focus on the single-agent cases~\cite{yang2017average, wang2017primal, chen2018scalable,Yang2016Efficient,Ghavamzadeh2007Hierarchical}. The research on average-reward MARL still undergoes exploration.
	
	
	There are two lines of research in existing literature that focus on the saddle-point formulation of RL. One line studies the saddle-point formulation resulted from the fixed-point problem of policy evaluation~\cite{6949624,dai2016learning,du2017stochastic,wai2018multi, lee2018primal}, i.e., learning the value function of a fixed policy. Among others, the works \cite{lee2018primal, wai2018multi} provided the sample complexity analysis of policy evaluation in the context of MARL, where the policies of all agents are fixed. 
	The other line, which includes this paper, focuses on the saddle-point formulation resulted from the policy optimization problem~\cite{wang2017primal,chen2018scalable}, where the policy is continuously updated towards the optimal one. This makes the analysis substantially more challenging than that for policy evaluation. In the single-agent setting, our work is closely related to~\cite{wang2017primal}. However, to the best of our knowledge, our work is the first to consider solving a saddle-point policy optimization in the context of MARL, which takes the coordination among multiple agents into account. Moreover, we also provide numerical simulations and case studies to corroborate our theoretical results, while previous works mainly focus on theoretical analysis~\cite{wang2017primal,chen2018scalable}.
	
	Finally, most MARL models are based on the parameter-consensus-based coordination, where the local agents consensus their parameters with a centralized server~\cite{lowe2017multi,8792117,Landon2016Multi,Nguyen2020Deep} or their neighboring agents~\cite{zhang2018fully, wai2018multi, 6949624, lee2018primal}.
	Although many works adopted voting-based coordination in their proposed learning algorithms~\cite{7738581,7786909,4407646}, they are not developed for MARL. A relevant work is~\cite{Partalas2007Multi}, which proposed a dedicated majority voting rule to coordinate the MARL agents under discounted MDP, which, however, is a heuristic strategy without theoretical guarantees and may not perform well on non-terminating tasks.
	
	
	\section{Problem Formulation}
	\label{sec:problem_formulation}
	In this paper, we consider the MARL in the presence of a generative model of the MDP~\cite{dietterich2013pac, taleghan2015pac, azar2013minimax}. The underlying MDP is unknown but having access to a sampling oracle, which takes an arbitrary state-action pair $ (i,a) $ as input and generates the next state $ j $ with probability $ p_{ij}(a) $, along with an immediate reward for each individual agent.
	The goal is to find the optimal policy of the unknown AMDP by interacting with the sampling oracle.
	Such a simulator-defined MDP has been studied by existing literatures in the context of single-agent RL, including the model-based RL~\cite{dietterich2013pac, taleghan2015pac, azar2013minimax} and model-free RL~\cite{kearns2002sparse, kearns1999finite}. In what follows, we first introduce the settings of the multi-agent AMDP and then formulate the multi-agent policy optimization problem as a primal-dual saddle point optimization problem.
	
	\subsection{Multi-Agent AMDP}
	We focus on the infinite-horizon AMDP, which aims at optimizing the average-per-time-step reward over an infinite decision sequence. 
	Existing works on RL usually impose a discount factor $ \gamma \in (0,1) $ on the future rewards to render the problem more tractable; e.g., by making the cumulative reward bounded. However, discounted RL may yield a poor performance over long-run (especially non-terminating) tasks and become computationally challenging when the discount factor is close to one~\cite{sutton2018reinforcement,yang2017average,Yang2016Efficient,Ghavamzadeh2007Hierarchical}. 
	In this paper, we do not assume that the future rewards are discounted. Rather, we assume that the AMDP satisfies certain fast mixing property~(given in Sec.~\ref{sec:theoretical}), which significantly complicates our analysis when compared with the discounted cases.
	
	A multi-agent AMDP can be described by the tuple
	\[ 
	\left( \mathcal{S}, \mathcal{A} , \mathcal{P}, \left\lbrace \mathcal{R}_m \right\rbrace^{M}_{m=1} \right),
	\] 
	where $ \mathcal{S} $ is the state space, $ \mathcal{A} $ is the action space, $ \mathcal{P} = \left\lbrace p_{ij}(a) \mid i, j \in \cS, a \in \cA \right\rbrace  $ is the collection of state-to-state transition probabilities, and $ \left\lbrace \mathcal{R}_m \right\rbrace^{M}_{m=1} $ is the collection of local reward functions with $\mathcal{R}_m = \left\lbrace r_{ij}^m(a) \mid i,j\in\cS, a\in\cA \right\rbrace$ and $M$ being the number of agents. We consider the setting where the reward functions of the agents may differ from each other and are private to each corresponding agent. We assume that the reward $r_{ij}^m(a)$, where $i,j\in\cS$, $a\in\cA$, and $m=1,\ldots,M$, lie in $[0,1]$. 
	This public state with private reward setting is widely considered in many recent works on collaborative MARL~\cite{wai2018multi,lee2018primal,zhang2018fully}.
	Moreover, we assume that the multi-agent AMDP is ergodic (i.e., aperiodic and recurrent), so that there is a unique stationary distribution under any stationary policy. 
	The MARL system selects the action to take according to the votes from local agents. Each agent determines its vote individually without communicating with others. In particular, at each time step $ t $, the MARL system works as follows:
	1) all agents observe the state $ i_t \in \mathcal{S} $;
	2) each agent votes for the action $ a_t $ to take under $ i_t $; 
	3) the system executes $ a_t $ according to the votes; 
	4) the system shifts to a new state $ i_{t+1} \in \mathcal{S} $ with probability $p_{i_ti_{t+1}}(a_t)$ and returns the rewards $ \lbrace r^m_{i_{t} i_{t+1} }(a_{t}) \rbrace^{M}_{m=1} $ to the agents.
	
	
	\subsection{Multi-Agent Policy Optimization}
	We denote the global acting policy, which determines the joint action to take, as $ \pi^g \in \Xi \subseteq \bbR^{|\cS| \times |\cA|} $, where $ \Xi $ consists of non-negative matrices whose $ (i,a) $-th entry $ \pi^g_{i,a} $ specifies the probability of taking action $ a $ in state $ i $.
	The multi-agent policy optimization problem aims at improving the global acting policy by maximizing the sum of local average-rewards, i.e.,
	\begin{equation}\label{eq:policy_optimization}
	\max_{\pi^g} \! \bigg\lbrace\! \bar{v}^{\pi^g} \!\!=\!\! \lim_{T\rightarrow\infty}\bbE^{\pi^g}  \!\bigg[ \frac{1}{T} \sum_{t=1}^{T} \! \sum_{m=1}^{M} r^m_{i_{t} i_{t+1} }(a_{t}) \Big\vert i_1 \!=\! i \bigg], i \in \cS  \bigg\rbrace\!,
	\end{equation}
	where $ \bbE^{\pi^g}[\cdot] $ denotes the expectation over all the state-action trajectories generated by the MARL system when following the acting policy $ \pi^g $.
	According to the theory of dynamic programming~\cite{puterman2014markov, bertsekas2005dynamic}, the value $ \bar{v}^* $ is the optimal average reward to problem~(\ref{eq:policy_optimization}) if and only if it satisfies the following Bellman equation: 
	\begin{equation}\label{eq:Bellman_equation}
	\begin{split}
	& \bar{v}^* + v^*(i) \\ 
	& = \max_{a\in\cA } \left\lbrace \sum_{j\in \cS} p_{ij}(a) v^*(j) \!+\! \sum_{j \in \cS}  p_{ij}(a) \!\! \sum_{m=1}^M r^m_{ij}(a) \!\right\rbrace, ~ \!\! \forall~ i \in \cS, 
	\end{split}
	\end{equation}
	where $ p_{ij}(a) $ is the transition probability from state $ i $ to state $ j $ after taking the action $ a $ and $ \boldsymbol{v}^* \in \mathbb{R}^{|\mathcal{S}|} $ is known as the difference-of-value vector that characterizes the transient effect of each initial state under the optimal policy~\cite{wang2017primal}. Note that there exist infinitely many $ \boldsymbol{v}^*$ that satisfy~\eqref{eq:Bellman_equation}; e.g., by adding constant shifts. However, this does not affect our analysis. More detailed descriptions of $ \boldsymbol{v}^*$ can be found in~\cite{wang2017primal}.
	
	\subsection{Saddle-Point Formulation}
	\label{sec:saddle_point_formularin}
	The Bellman equation in (\ref{eq:Bellman_equation}) can be written as the following linear programming problem:
	\begin{equation}\label{eq:primal}
	\begin{aligned}
	\min_{\bar{v},\boldsymbol{v}} & \quad  \bar{v} \\
	\mathrm{s.t.} & \quad \bar{v} \cdot \boldsymbol{e} + \left( I -  P_a \right)  \boldsymbol{v} -  \sum_{m=1}^M \bar{\boldsymbol{r}}^m_a  \geq \bm{0}, \ \ \forall~a\in \mathcal{A},
	\end{aligned}
	\end{equation}
	where $ P_a \in \mathbb{R}^{|\cS| \times |\cS|} $ is the MDP transition matrix under action $ a $ whose $ (i,j) $-th entry is $ p_{ij}(a) $ and $ \bar{\boldsymbol{r}}^m_a \in \mathbb{R}^{|\cS|} $ is the expected state-transition reward under action $ a $ with $ \bar{r}^{m}_{i,a} = \sum_{j \in \cS} p_{ij}(a) r^m_{ij}(a), \ \forall i \in \cS $.
	The dual of (\ref{eq:primal}) can be written as 
	\begin{equation}\label{eq:dual}
	\begin{aligned}
	\max_{\bmu} & \quad \sumi \suma \sum_{m=1}^M \mu_{i,a} \bar{r}^m_{i,a} \\
	\mathrm{s.t.} & \quad \sum_{a\in \mathcal{A}} \bmu_a^{\top} \left({I}  - P_a\right) = \bm{0}, \\
	& \quad \sum_{i \in \mathcal{S}} \sum_{a \in \mathcal{A}} \mu_{i, a} = 1, \ \mu_{i, a} \geq 0,
	\end{aligned}
	\end{equation}
	where $ \bmu $ is the dual variable. By linear programming strong duality, if $(\bar{v}^*,\bm{v}^*)$ and $\bm{\mu}^*$ are optimal solutions to the primal and dual problems~\eqref{eq:primal} and~\eqref{eq:dual}, respectively, then they satisfy the zero complementarity gap condition:
	\begin{equation} \label{eq:opt-cond}
	\begin{split}
	0 &= \sum_{a\in\cA} (\bm{\mu}_a^*)^\top \left( \bar{v}^* \cdot \boldsymbol{e} + \left( I -  P_a \right)  \boldsymbol{v}^* -  \sum_{m=1}^M \bar{\boldsymbol{r}}^m_a  \right) \\
	&= \bar{v}^* + \sum_{a\in\cA} (\bm{\mu}_a^*)^\top \left( \left( I -  P_a \right)  \boldsymbol{v}^* -  \sum_{m=1}^M \bar{\boldsymbol{r}}^m_a  \right).
	\end{split}
	\end{equation}

	Observe that problems~\eqref{eq:primal} and~\eqref{eq:dual} involve rather complicated constraints. Hence, it is common to consider their saddle-point formulation, whose constraints are simpler:
	\begin{equation}\label{eq:saddel-point}
	\min_{\boldsymbol{v} \in \cV} \max_{\bmu \in \cU} \ \sum_{a\in \mathcal{A}} \bmu_a^{\top} \left( ( P_a - I ) \boldsymbol{v} + \sum_{m=1}^M \bar{\boldsymbol{r}}^m_a \right).
	\end{equation}
	Here, 
	\[ \cV=\mathbb{R}^{|\cS|}, \ \cU=\left\{\bm{\mu}\in\mathbb{R}^{|\cS|\times|\cA|} \,\Big|\, \sum_{i \in \mathcal{S}} \sum_{a \in \mathcal{A}} \mu_{i, a} = 1, \ \bm{\mu} \ge \bm{0} \right\}\]
	are the primal and dual constraint sets, respectively. Later, we shall focus on multi-agent AMDPs that satisfy certain fast mixing property. This will allow us to use a smaller but still structured primal constraint set $\cV$; see Sec.~\ref{sec:theoretical}.
	
	
	It is known that there is a correspondence between randomized stationary policies and feasible solutions to the dual problem~\eqref{eq:dual}~\cite{puterman2014markov}. In particular, given an optimal dual solution $ \bmu^* \in \mathbb{R}^{|\cS|\times |\cA|} $, the optimal acting policy $ \pi^g $ can be obtained via $ \pi^*_{i,a} = \mu^*_{i,a} / \sum_{a\in\mathcal{A}} \mu_{i,a}^*$. Hence, our goal now is to obtain an optimal dual solution $ \bmu^*$. 
	
	\section{Voting-Based Learning Algorithm}
	\label{sec:algorithm}
	In this section, we propose a voting mechanism that specifies how local votes determine the global action. Then, we prove that the voting mechanism yields an equivalence between the update on the global acting policy and that on the distributed voting policies. Consequently, problem~(\ref{eq:saddel-point}) can be solved in a distributed manner, and we propose a primal-dual learning algorithm for it.
	
	\subsection{Voting Mechanism}
	We denote the pair of primal and dual variables corresponding to the global acting policy $ \pi^g $ as $ \bv^g $ and $ \bmu^g $, respectively. We also introduce a pair of local primal and dual variables corresponding to each local voting $ \pi^m$ ($m=1,\ldots,M$) as $ \bv^m $ and $ \bmu^m $, where $ \pi^m \in \Xi \subseteq \bbR^{|\cS| \times |\cA|} $ is a randomized stationary policy. Then, the voting mechanism takes the form
	\[ 
	\mu^{g,t}_{i,a} \propto \prod_{m=1}^{M} \mu^{m, t}_{i,a}.
	\] 
	The voting mechanism indeed reveals the relationship between the global acting policy and the local voting policies. 
	
	\subsection{Primal-Dual Learning Algorithm}
	We now develop a primal-dual learning algorithm to solve problem~(\ref{eq:saddel-point}) in a distributed manner based on a double-sampling strategy. 
	Recall that we consider the MARL under a generative MDP, where the agents are interacting with a black-box sampling oracle to learn the optimal policy. The sampling oracle works in a similar way as the experience replay used in deep RL models~\cite{lowe2017multi,8792117,foerster2017stabilising,xu2019load}. In practical applications, the sampling oracle or experience replay can be placed in a centralized node which can communicate with the local agents, as in many existing MARL frameworks~\cite{lowe2017multi,8792117,Landon2016Multi,Nguyen2020Deep}. However, it only needs to collect the vote information $ \mu^m_{i,a} $ from the agents in order to coordinate the sampling during the learning process, instead of performing iterative parameter consensus as existing methods~[14], [18], [29], [30].
	The detailed procedure is provided in Algorithm~\ref{algorithm:MARL}. 
	In what follows, we first introduce the local dual and primal updates in our algorithm. Then, we prove that the local updates are equivalent to the global updates if the voting mechanism is specified properly.
	
	\subsubsection{Local Dual Update}
	We update the local dual variables based on uniform sampling. 
	Specifically, the first state-action pair $ \left(i_t, a_t\right) $  to update the local dual variables is sampled with uniform probability $ p^{\text{dual}}_{i,a} = \frac{1}{|\mathcal{S}|\cdot|\mathcal{A}|} $. 
	The MARL system then shifts to the next state $ j_t $ conditioned on $ (i_t, a_t) $ and returns the local rewards $ \lbrace r^m_{i_{t}j_t }(a_{t}) \rbrace^{M}_{m=1} $ to the agents.
	The local dual variable $ \bmu^{m,t} $ of agent $ m $ is updated as 
	\begin{equation} \label{eq:local-dual-update}
	\mu^{m, t+1}_{i,a} \!=\! \left\{ \begin{array}{ll} \mu^{m, t}_{i,a} \exp\big\lbrace \Delta^{m, t}_{i,a} \big\rbrace, & \text{if}~ i = i_t, \, a = a_t, \\ \mu^{m, t}_{i,a}, & \text{otherwise}, \end{array} \right.
	\end{equation}
	where 
	\begin{equation}\label{eq:local-dual-gradient}
	\Delta^{m, t}_{i,a} = \beta \left( \frac{\frac{1}{\beta}\log x^t + v_j^t - v_i^t - C}{M} + r^m_{ij}(a) \right)
	\end{equation}
	with $(i,a,j)=(i_t,a_t,j_t)$, $\beta > 0$ being the step-size, $C$ being a parameter to be specfied, and
	\begin{equation}\label{eq:local-xt}
	x^{t} = \frac{1}{\sum_{i \in \mathcal{S}, a \in \mathcal{A}} \prod_{m=1}^{M} \mu^{m, t}_{i,a}}.
	\end{equation}
	Here, $ x^t $ can be viewed as the proportion between the locally recovered partial derivatives and the global true partial derivatives of the minimax objective in (\ref{eq:saddel-point}). It also defines the explicit form of the voting mechanism; see Lemma~\ref{lemma:equivalent-global} below. 
	However, it is important to note that we do not need to compute $x^t$ in our algorithm, as it does not influence the sampling in the subsequent primal update step and is used purely for analysis purposes. In other words, one can remove the term of $ \log x^t $  from (\ref{eq:local-dual-gradient}) without influencing the learning performance.
	

	
	\subsubsection{Local Primal Update}
	We update the local primal variables based on probability sampling, where the probability is specified by the dual variables. 
	Specifically, the second state-action pair $ \left(i_t, a_t\right) $ to update the local primal variables is sampled with probability 
	\begin{equation}\label{eq:prim-sampl-prob}
	p^{\text{primal}}_{i_t,a_t} = \frac{\prod_{m=1}^{M} \mu^{m, t}_{i_t,a_t}}{\sum_{i \in \mathcal{S}, a \in \mathcal{A}} \prod_{m=1}^{M} \mu^{m, t}_{i,a}}.
	\end{equation}
	The system then shifts to the next state $ j_t $ conditioned on $ (i_t, a_t) $, and returns the local rewards to the agents.
	The local primal variable $ \bv^t $ is updated as
	\begin{equation}  \label{eq:local-prim-up}
	\bv^{t+1} = \Pi_{\mathcal{V}} \left\lbrace \bv^{t} +\bd^{t} \right\rbrace,
	\end{equation}
	where 
	\begin{equation} \label{key-1}
	\bd^{t} = \alpha(\bm{e}_i-\bm{e}_j)
	\end{equation}
	with $(i,j)=(i_t,j_t)$; $ \alpha > 0$ is the step-size; $ \Pi_{\mathcal{V}} \left\lbrace \cdot \right\rbrace $ denotes the projector onto the search space $ \mathcal{V} $, which will be defined in Sec.~\ref{sec:theoretical}.
	Note that the local primal update is identical across the agents. Hence, we use the same notation $ v^t_i $ in the primal update for all the agents in the sequel.
	
	\subsubsection{Communication}
	The centralized sampling oracle needs to collect the vote information $ \mu^m_{i,a} $ to compute the probability $ p^{\text{primal}}_{i_t,a_t} $ according to (\ref{eq:prim-sampl-prob}) and returns the reward information to the agents. However, note that the vote information $ \mu^m_{i,a} $ and the reward information $ r^m_{i,j}(a) $ of each agent is a scalar, such that the communication overhead at each learning step of our method only scales as $ \mathcal{O}(M) $.
	In contrast, most existing MARL methods are developed based on parameter consensus, where local agents need to reach consensus on its value or policy function with a centralized center~\cite{lowe2017multi,8792117,Landon2016Multi,Nguyen2020Deep} or their nearby agents~\cite{zhang2018fully, wai2018multi, 6949624, lee2018primal}. Since the value or policy function scales as $ \mathcal{O}(|\mathcal{S}|\cdot|\mathcal{A}|) $, the communication overhead at each learning step of their models scales as $ \mathcal{O}(M \cdot |\mathcal{S}|\cdot|\mathcal{A}|) $. Although this cost can be reduced if they adopt linear or nonlinear function to approximate the value or policy function, it is still related to the size of the function approximators, which can be enormous if they are deep neural networks. Moreover, the exchanged information in our algorithm is the vote information, instead of the privacy-sensitive policy or value information, which can alleviate privacy and security concerns considerably.
	
	\subsubsection{Equivalent Global Update}
	We now prove that with a properly specified voting mechanism, the primal-dual updates on the local voting policies are equivalent to the centralized primal-dual updates on the global acting policy.
	
	\begin{Lemma}[Equivalent Global Update] \label{lemma:equivalent-global}
		By specifying the voting mechanism as 
		\begin{equation}\label{eq:global-dual-end}
		\mu^{g, t}_{i,a} = x^{t} \prod_{m=1}^{M} \mu^{m, t}_{i,a},
		\end{equation}
		where $ x^{t}$ is given by~\eqref{eq:local-xt}, the local primal-dual updates~\eqref{eq:local-dual-update} and~\eqref{eq:local-prim-up} are equivalent to the following global primal-dual updates:
		\begin{subequations} \label{eq:glob-up}
			\begin{align}
			\mu^{g, t+1}_{i,a} &= x^{t+1}\mu^{g, t}_{i,a} \exp\left\lbrace {\Delta}^{g, t}_{i,a} \right\rbrace, \ \forall \ i \in \cS, \ a \in \cA, \label{eq:glob-1} \\
			\boldsymbol{v}^{t+1} & = \Pi_{\mathcal{V}} \left\lbrace\boldsymbol{v}^t + \boldsymbol{d}^{t} \right\rbrace. \label{eq:glob-2}
			\end{align}
		\end{subequations}
		Here,
		\begin{equation} \label{key-0}
		\Delta^{g, t}_{i,a} = \beta \left(  v_j ^t -  v_i^t -  C  +  \sum_{m=1}^M r^m_{ij}(a) \right)
		\end{equation}
		and $\bm{d}^{t}=\alpha(\bm{e}_i-\bm{e}_j)$, where $(i,a)=(i_t,a_t)$ with probability $\mu^{g,t}_{i_t,a_t}$ and $j=j_t$ is obtained from the system by conditioning on $(i_t,a_t)$. \hfill$\blacksquare$
	\end{Lemma}
	
	We remark that the global primal-dual updates~\eqref{eq:glob-up} are conditionally unbiased partial derivatives of the minimax objective given in~(\ref{eq:saddel-point}).
	
	{\smallskip\noindent\it Proof.}
	Recall that the local dual variable $ \bmu^{m,t} $ of agent $ m $ is updated by~\eqref{eq:local-dual-update}. We now prove a recursive relationship between $ \mu^{g, t+1}_{i,a} $ and $ \mu^{g, t}_{i,a} $ as follows.
	Given $ (i,a) = (i_t,a_t) $, starting from the voting mechanism defined in~\eqref{eq:global-dual-end}, we have
	\begin{align*}
	\mu^{g, t+1}_{i,a} & = x^{t+1} \prod_{m=1}^{M} \mu^{m, t+1}_{i,a} \nonumber \\
	&= x^{t+1} \prod_{m=1}^{M} \Big(\mu^{m, t}_{i,a} \exp\big\lbrace \Delta^{m, t}_{i,a} \big\rbrace \Big) \nonumber \\
	&= x^{t+1} \prod_{m=1}^{M} \mu^{m, t}_{i,a} \exp \left\lbrace \sum_{m=1}^M \Delta^{m, t}_{i,a} \right\rbrace \\
	&= x^{t+1} (x^{t})^{-1} \mu^{g, t}_{i,a} \exp \left\lbrace \sum_{m=1}^M \Delta^{m, t}_{i,a} \right\rbrace \\
	&= x^{t+1} \mu^{g, t}_{i,a} \exp \left\lbrace \beta \left(  v_j^t  - v_i^t  - C + \sum_{m=1}^M r^m_{ij}(a) \right) \right\rbrace. \nonumber
	\end{align*}
	Hence, using the definition of $\Delta^{g, t}_{i,a}$ in~\eqref{key-0}, the local dual update based on $ \Delta^{m, t}_{i,a} $ can be equivalently expressed as the global dual update based on $ \Delta^{g, t}_{i,a} $, i.e.,~\eqref{eq:glob-1} holds.
	
	As for the local primal update, since the oracle generates the second sample with probability $p^{\text{primal}}_{i,a}$ given by~\eqref{eq:prim-sampl-prob}, which is exactly the same as the global dual variable $ \mu^{g, t}_{i,a} $ given in~\eqref{eq:global-dual-end}, the local and global primal updates are identical. \hfill $\blacksquare$
	
	\medskip
	\begin{Lemma}[Unbiasedness] \label{lemma:unbiasedness}
		Consider the voting mechanism in Lemma~\ref{lemma:equivalent-global}. Let $\mathcal{F}_t$ be the filtration at time $t$, i.e., information about all the state-action pair sampling and state transition right before time $t$. Then, the dual update weight $ \Delta^{g, t}_{i,a} $
		is, up to a constant shift, a multiple of the conditional partial derivative of the minimax objective in~\eqref{eq:saddel-point} with respect to $\mu_{i,a}$:
		\begin{align*}
		& \mathbb{E} \big[ \Delta^{g, t}_{i,a} \mid  \mathcal{F}_t \big] \\
		& = \frac{\beta}{|\cS|\cdot|\cA|} \left( (P_a - I) \boldsymbol{v}^{t} + \sum_{m=1}^M \bar{\boldsymbol{r}}^m_a - C \cdot \bm{e} \right)_i, \\
		&\qquad\qquad \forall \ i\in\cS, \ a \in \cA.
		\end{align*}
		Moreover, the primal update weight $ d_i^{t} $ is a multiple of the conditional partial derivative of the minimax objective in~\eqref{eq:saddel-point} with respect to $v_i$:
		\[ 
		\mathbb{E} \left[ \bm{d}^{t} \mid  \mathcal{F}_t \right]  = \alpha \suma (I - P_a)^\top \bm{\mu}_a^{g,t}. 
		\] 
		\hfill$\blacksquare$
	\end{Lemma}
	{\smallskip\noindent\it Proof.}
	For arbitrary $i\in\cS$ and $a\in\cA$, we use~\eqref{key-0} to compute
	\begin{align*}
	& \frac{1}\beta\cdot \bbE \left[  \Delta^{g, t}_{{i,a}}  \mid \cF_t \right]   \\
	&\begin{aligned}
	&= \frac{1}{|\cS|\cdot|\cA|}  \left( \sumj p_{ij}(a)v^t_j - v^t_{i} \right) \\
	&\qquad+\frac{1}{|\cS|\cdot|\cA|} \left( \sumj \sum_{m=1}^M p_{ij}(a) r^m_{ij}(a) - C \right)
	\end{aligned}\\
	& = \frac{1}{|\cS|\cdot|\cA|} \left( (P_a - I) \bv^t + \sum_{m=1}^M \bar{\boldsymbol{r}}^m_a - C \cdot \bm{e} \right)_i.
	\end{align*}
	
	On the other hand, using~\eqref{key-1} and the fact that the state-action pair for updating the primal variables is generated with probability $\bm{\mu}^{g,t}$, we compute, for an arbitrary $i\in\cS$,
	\begin{align*}
	& \bbE \left[ \bm{d}^{t} \mid \cF_t \right]   \\
	&=  \alpha\left[ \sum_{i\in\cS} \Pr \left( i_t = i \mid \cF_t \right) \bm{e}_i - \sum_{j\in\cS}  \Pr \left( j_t = j \mid \cF_t \right) \bm{e}_j \right] \\
	&= \alpha \left[ \sum_{i\in\cS} \suma \mu^{g,t}_{i,a} \bm{e}_i - \sumj \sumi \suma p_{ij}(a) \mu^{g,t}_{i,a} \bm{e}_j \right] \\
	&=  \alpha \suma (I - P_a)^\top \bm{\mu}_a^{g,t}.
	\end{align*}
	This completes the proof. \hfill $\blacksquare$
	
	\begin{algorithm}[!h]
		\caption{Voting-Based MARL}
		\label{algorithm:MARL}
		\begin{algorithmic}[1]
			\STATE \textbf{Initialization:} MARL tuple $\mathcal{M}=(\mathcal{S}, \mathcal{A}, \mathcal{P}, \left\lbrace \mathcal{R}_m \right\rbrace^{M}_{m=1} )$, time horizon $T$, parameters
			\begin{align*}
			\alpha &= (4\tmix+M) \sqrt{ \frac{|\cS|}{|\cA|} \cdot \frac{\log (|\cS|\cdot|\cA|) }{2T}}, \\
			\beta  &= \frac{1}{4\tmix+M} \sqrt{ \frac{|\cS|\cdot|\cA|\cdot\log (|\cS|\cdot|\cA|) }{2T}}, \\
			C &= 4t^{*}_{\text{mix}} + M.
			\end{align*}
			\STATE{Set $ \boldsymbol{v} = \boldsymbol{0} \in \mathbb{R}^{|\mathcal{S}|} $ and $ \mu^{m, 0}_{i,a} = \frac{1}{|\mathcal{S}|\cdot|\mathcal{A}|}, \ \forall \ i \in \mathcal{S}, \  a \in \mathcal{A} $.}
			\STATE \textbf{Iteration:}
			\FOR{$ t = 1, 2, \ldots, T $}
			\STATE{The system samples $ (i_t , a_t) $ with probability $ p^{\text{dual}}_{i_t,a_t} $.}
			\STATE{The system shifts to next state $ j_t $ conditioned on $ (i_t, a_t) $ and generates the local rewards $ \left\lbrace r^m_{i_tj_t}(a_t) \right\rbrace^{M}_{m=1}  $.}
			\FOR{ $ m = 1, 2, \ldots, M $ }
			\STATE{The agent $ m $ updates its local dual variable according to~\eqref{eq:local-dual-update}.}
			\ENDFOR
			\STATE{The system collects the updated $ \mu^{m,t+1}_{i,a} $ from local agents and samples $ (i_t , a_t) $ with probability $ p^{\text{primal}}_{i_t,a_t}$.}
			\STATE{The system shifts to next state $ j_t $ conditioned on $ (i_t, a_t) $ and generates the local rewards $ \left\lbrace r^m_{i_tj_t}(a_t) \right\rbrace^{M}_{m=1}  $which are returned to the agents.}
			\FOR{ $ m = 1, 2, \ldots, M $}
			\STATE{The agent $ m $ updates its local primal variable according to~\eqref{eq:local-prim-up}.}
			\ENDFOR
			\ENDFOR
			\STATE{Set $ \hat{\mu}^g_{i,a} = \frac{1}{T} \sum_{t=1}^{T}  \prod_{m=1}^{M} \mu^{m, t}_{i,a} , ~\forall \ i \in \mathcal{S}, \  a \in \mathcal{A} $.}
			\STATE{\textbf{Return:}} $ \hat{\pi}^g_{i, a} = \frac{\hat{\mu}^g_{i,a}}{\sum_{a \in \mathcal{A}} \hat{\mu}^g_{i,a}}, \ \forall \ i \in \mathcal{S}, \  a \in \mathcal{A} $.
		\end{algorithmic}
	\end{algorithm}
	
	\section{Theoretical Results}
	\label{sec:theoretical}
	In this section, we present the convergence analysis of Algorithm~\ref{algorithm:MARL}. 
	We start by making the following assumption on the considered multi-agent AMDP. A similar assumption has also been used in~\cite{chen2018scalable, wang2017primal} for the case of a single-agent RL.
	
	%
	\begin{Assumption} \label{assumption:mix}
		There exists a constant $ t_{\text{mix}}^* > 0 $ such that for any stationary policy $ \pi^g $, we have
		\[ 
		t_{\text{mix}}^* \geq \min_{t} \left\lbrace t \,\Big|\, \| (P^{\pi^g})^{t}(i,\cdot) - \bnu^{\pi^g} \|_{TV} \leq \frac14, ~\forall \ i \in \cS \right\rbrace, 
		\] 
		where $\| \cdot\|_{TV}$ is the total variation and $P^{\pi^g}(i,j) = \sum_{a\in\mathcal{A}} \pi_{i,a}^gp_{ij}(a)$. \hfill$\blacksquare$
	\end{Assumption}
	
	\smallskip
	\noindent The above assumption requires the multi-agent AMDP to be sufficiently rapidly mixing, with the parameter $ t_{\text{mix}}^* $ characterizing how fast the multi-agent AMDP reaches its stationary distribution from any state under any acting policy~\cite{wang2017primal}. 
	In particular, $ t_{\text{mix}}^* $ controls the distance between any stationary policy and the optimal policy under the considered multi-agent AMDP.
	It has been shown in~\cite{wang2017primal} that under Assumption~\ref{assumption:mix}, an optimal difference-of-value vector $\bm{v}^*$ satisfying $ \norm{\bm{v}}_{\infty} \leq 2 t_{\text{mix}}^*$ exists.
	
	Based on the above discussion, we can use the following smaller constraint set $\cV$ for the global primal variable $ \bv $:
	\[ \mathcal{V} = \left\lbrace \bm{v} \in \mathbb{R}^{|\mathcal{S}|} \,\Big|\, \| \boldsymbol{v} \|_{\infty} \leq 2 t^*_{\text{mix}} \right\rbrace. \]
	%
	%
	Now, we are ready to establish the convergence of our proposed Algorithm~\ref{algorithm:MARL}.
	\begin{Theorem}[Finite-Iteration Duality Gap] \label{thm:duality-gap}
		Let $\mathcal{M}=(\cS,\cA,\cP,\left\lbrace \mathcal{R}_m \right\rbrace^{M}_{m=1}  )$ be an arbitrary multi-agent AMDP tuple satisfying Assumption \ref{assumption:mix}. Then, the sequence of iterates generated by Algorithm \ref{algorithm:MARL} satisfies
		\begin{align*}
		& \bar v^* + \frac1T\sum^T_{t=1}\bbE \left[ \suma \left( (I -  P_{a}) \bv^* - \sum_{m=1}^M \bar{\boldsymbol{r}}^m_{a} \right)^{\top} \bm{\mu}^{g,t}_{{a}} \right] \\ 
		& \leq \tilde{O} \bigg( ( 4\tmix+M)  \sqrt{\frac{ |\cS|\cdot|\cA| }{T}} \bigg),
		\end{align*}
		where $\tilde{O}(\cdot)$ hides polylogarithmic factors. 	\hfill$\blacksquare$
	\end{Theorem}
	
	\smallskip
	\noindent Recall from~\eqref{eq:opt-cond} that the complementarity gap of a pair of optimal solutions to the primal-dual problems~\eqref{eq:primal} and~\eqref{eq:dual} is zero. Hence, 	Theorem~\ref{thm:duality-gap} suggests that the iterates $\{\bm{\mu}^{g,t}\}_{t\ge0}$ converge to an optimal solution to the dual problem~\eqref{eq:dual} at a sublinear rate. The result also covers the single-agent RL~\cite{wang2017primal} as a special case, which makes our model more general. We defer the proof of Theorem~\ref{thm:duality-gap} to the appendix.
	
	It is worth pointing out that in our proof, the scalar $ M $ in Theorem~\ref{thm:duality-gap}, i.e., the number of agents, comes from the bound of the total reward of all agents $ \sum_{m=1}^M r_{ij}^m(a) \in [0, M] $, $\forall \ i,j \in \cS, \ a \in \cA$. As such, if we consider a normalized reward where $ \sum_{m=1}^M r_{ij}^m(a) \in [0, 1] $, then the complexity in Theorem~\ref{thm:duality-gap} will be independent of $ M $.

	\section{Numerical Results}
	\label{sec:simulation}
	In this section, we evaluate the proposed voting-based MARL algorithm through two case studies. 
	In the first case study, we verify the convergence of our proposed algorithm with the generated MDP instances. 
	In the second case study, we exhibit how to apply our proposed algorithm to  solve the placement optimization task in a UAV-assisted IoT network, where the ground base stations and the UAV-mounted base station are treated as the IoT devices. The UAV-mounted base station collects vote information from the ground base stations, which is then used to determine the placement of the the UAV-mounted base station to maximize the overall system capacity.
	Our results show that the distributed decision making does not slow down the process of achieving global consensus on optimality and that voting-based learning is more efficient than letting agents behave individually and selfishly. 
	
	\subsection{Empirical Convergence}
	\begin{figure}[tb]
		\centering
		\subfigure[Convergence of duality gap]
		{ 	
			\label{fig:v}
			\includegraphics[trim = 5 5 5 5, clip, width=0.8\columnwidth]{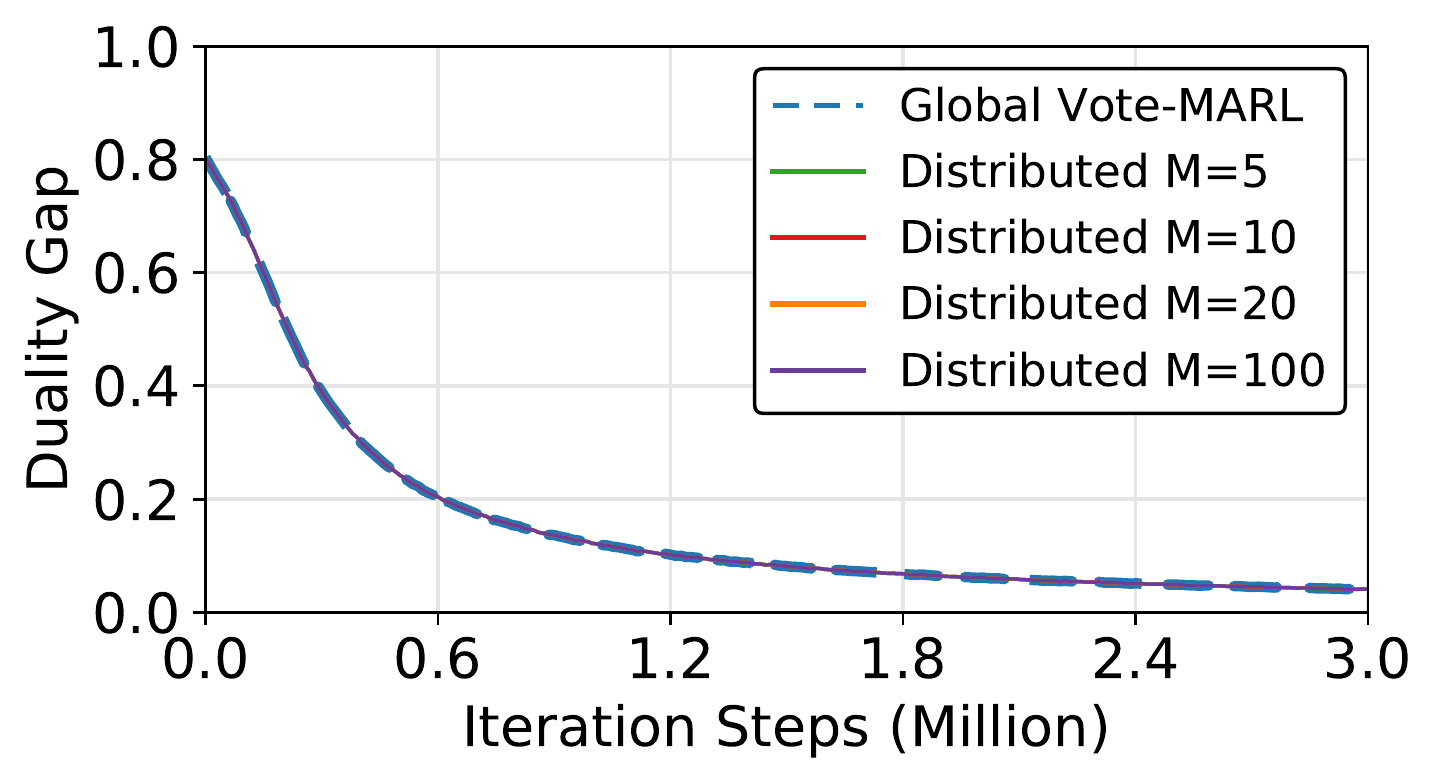}
		}
		\subfigure[Convergence to the global optimal policy]
		{ 	
			\label{fig:pi}
			\includegraphics[trim = 5 5 5 5, clip, width=0.8\columnwidth]{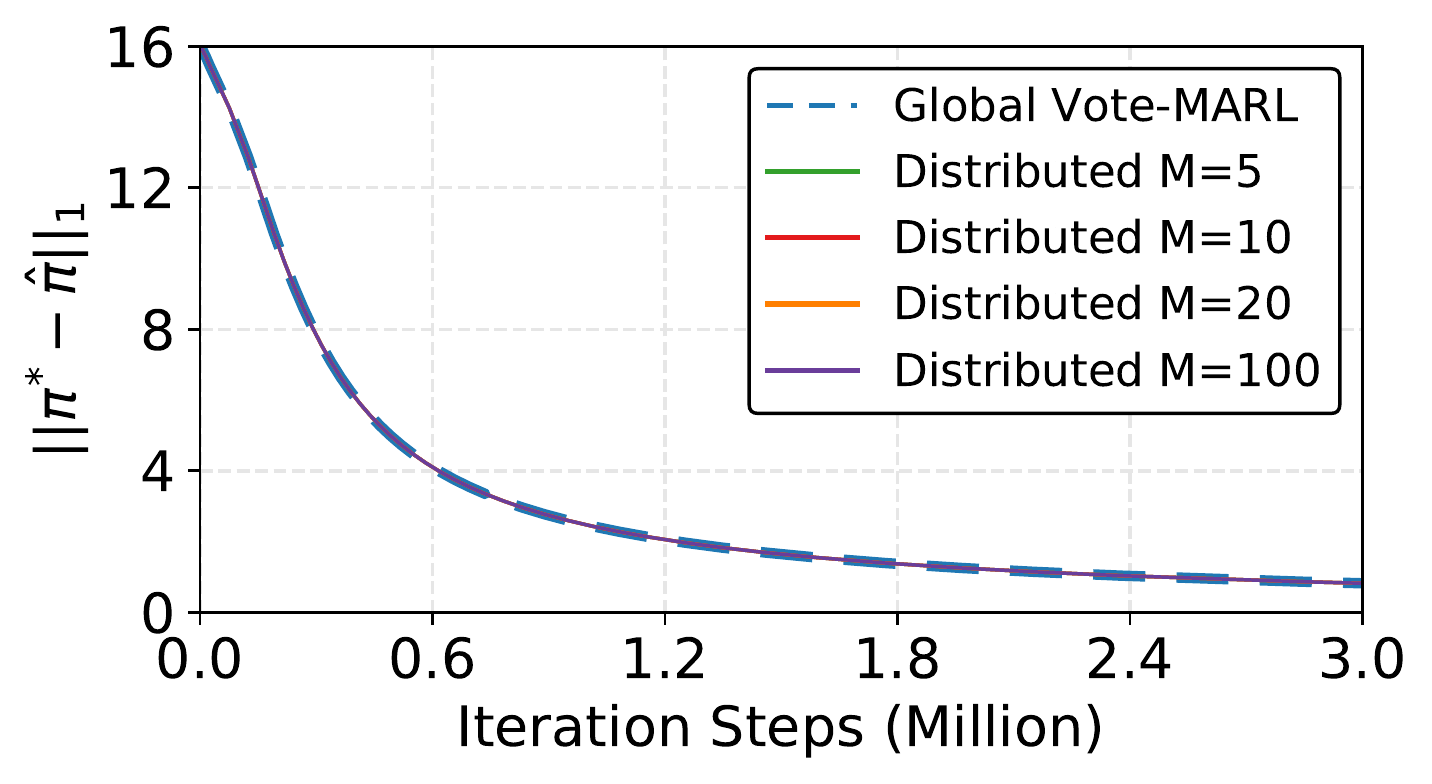}
		}
		\caption{Empirical convergence rate of the proposed algorithm. Each generated MDP instance contains $ |\cS| = 50 $ states and $ |\cA| = 10 $ actions at each state. The number of local agents varies from $ M=5 $ to $ M=100 $ with the total reward of all agents bounded in $ [0, 1] $.}
		\label{fig:error}
	\end{figure}
	We generate instances of the multi-agent MDP using a similar setup as in~\cite{Adam2016}. 
	Specifically, given a state and an action, the multi-agent MDP shifts to the next state assigned from the entire set without replacement. 
	The transition probabilities are generated randomly from $ [0,1] $ and then normalized so that they sum to one.
	The optimal policy is generated with purposeful behavior by letting the agent favor a single action in each state and assigning it with a higher expected reward in $ [0, 1] $. 
	
	In Fig.~\ref{fig:error}, we show the empirical convergence results of 
	1) the duality gap, i.e., the one given in Theorem~\ref{thm:duality-gap};
	2) the distance between the optimal policy and the learned policy, i.e., $ \norm{\boldsymbol{\pi}^{*} - \boldsymbol{\hat{\pi}}}_1 $.
	The convergence curves are averaged over $ 100 $ instances.
	Generally, the empirical convergence rates corroborate the result given in Theorem~\ref{thm:duality-gap}.
	Besides, we also present 1) the performance change as the number of local agents varies from $ M=5 $ to $ M=100 $ and 2) the performance of centralized learning, which directly uses the global primal-dual updates to learn the global policy. The result shows that the empirical convergence rates of the centralized case and the distributed case are the same for different numbers of agents $ M $. This indicates that distributed decision making does not slow down the process of achieving global consensus on optimality.
	
	\subsection{Application in Multi-Agent IoT Systems}
	We now apply the proposed voting-based MARL algorithm to a multi-agent IoT system which contains ground base stations, smartphones, and UAVs.
	In particular, UAV-assisted wireless communication has recently attracted much attention~\cite{mozaffari2018tutorial, 8377340, 7762053, 7010528},
	due to that UAV mounted with a mobile base station (UAV-BS) can provide high-speed air-to-ground data access by using the line-of-sight (LoS) communication links.
	However, obtaining the best performance in an UAV-BS-assisted wireless system highly depends on the placement of the UAV-BS~\cite{mozaffari2018tutorial, 8377340, 7762053}. 
	Here, we consider optimizing the placement of UAV-BS continuously through our proposed voting-based MARL algorithm.
	
	Existing works on the placement optimization of UAV-BS have two major drawbacks. First, many of them do not consider user movements~\cite{8423028, 7918510, 7762053, 7510820, 7122576}, but the change of user distribution can largely influence the system performance.
	Second, many of them determine the optimal placement of UAV-BS by assuming that the performance gain of each ground BS is public information~\cite{8377340, 7122576}, which may be impractical in real-world wireless systems that have mixed wireless operators, infrastructures, and protocols.
	To overcome these drawbacks, we model the UAV-BS placement optimization as a voting-based MARL problem, where multiple ground BS learn to place the UAV-BS optimally with adaptation to user movements and without the need to share their reward information. The aim is to maximize the global performance gain of all ground BS.
	
	\begin{figure}[tb]
		\centering
		\includegraphics[trim = 1 1 1 1, clip, width=0.8\columnwidth]{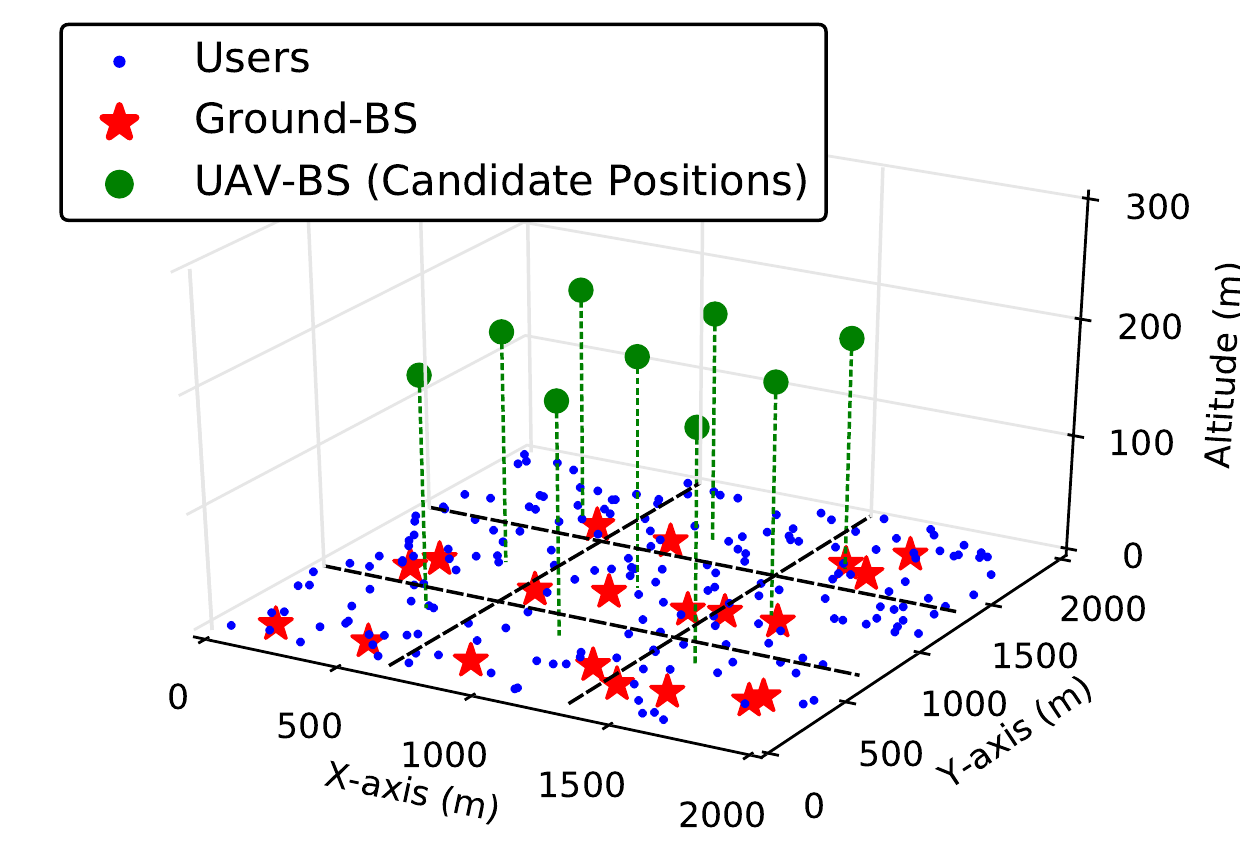}
		\caption{3D distribution of the investigated $ 4 $km$^2 $ area, which contains $ M = 20 $ randomly deployed ground BS, one UAV-BS flying among $ |\cA| = 9 $ candidate aerial locations, and $ 200 $ mobile users moving by following the random walk model~\cite{camp2002survey}. The action is to move the UAV-BS to any one of the $ |\cA| = 9 $ candidate aerial locations, which is determined by the votes from the ground BS.}
		\label{fig:scenario}
	\end{figure}
	
	We consider the downlink of a wireless cellular network. As shown in Fig.~\ref{fig:scenario}, the 2km$\times$2km area of interest has $ M = 20 $ regularly deployed ground BS, one UAV-BS flying at $200$m to provide air-to-ground communications, and $ 200 $ mobile users moving according to the random walk model in~\cite{camp2002survey}, each having a constant-bit-rate communication demand. 
	The UAV-BS can move to any one of the aerial locations from a finite set $ |\mathcal{A}| $ to provide air-to-ground communication. The user mobility follows the random walk model in~\cite{camp2002survey}, where each user moves at an angle uniformly distributed between $ [0, 2\pi] $ and a random speed between $ [0, c_{\text{max}}] $ with $ c_{\text{max}} $ being the maximum moving speed. 
	Table~\ref{table:parameters} summarizes the main parameters. 
	The air-to-ground channel and ground-to-ground channel are modeled according to~\cite{Challita2019Interference}~(Sec.~II). The load of each base station is defined as the ratio between the required number of PRBs and the total number of available PRBs according to~\cite{xu2019load}~(Sec.~II-B).
	\begin{table}[t]
		\centering
		\caption{Parameters}
		\label{table:parameters}
		\scalebox{1}{
			\begin{tabular}{@{}ll@{}}
				\toprule
				\textbf{Parameters} & \textbf{Values} \\ \midrule
				CBR ($ C_u $)                            & $ 128 $ kbps \\
				Total 2D area                            &  $ 4 $ km$^2 $ \\
				Total bandwidth                          & $ 20 $ MHz \\
				Carrier frequency ($ f_c $)              & $ 2 $ GHz \\
				PRB bandwith ($ B $)                     & $ 180 $ kHz \\
				Max user velocity ($ c_{\text{max}} $)          & $ 10 $ m/s \\
				Ground BS max transmit power ($ P_m $)   & $ 46 $ dBm \\
				UAV-BS max transmit power ($ P_U $)      & $ 20 $ dBm \\ 
				Additional LoS path loss ($ \eta_{LoS} $) & $ 1 $ dB \\
				Noise power spectral density ($ N_0 $)   & $ -174 $ dBm/Hz \\
				\bottomrule
			\end{tabular}
		}
	\end{table}

	The learning context is defined as follows.
	1)~States: We divide the area of interest into $ 3 \times 3 $ grids and use the load of each grid to characterize the wireless system status. The load of each grid is indicated by one of two states: a) overloaded, if the users' demand within the grid is higher than the mean demands of all the grids; b) underloaded, otherwise. Since the grids cannot be all overloaded or all underloaded, there are only $ |\cS| = 510 $ states for the wireless system with $ 9 $~grids.
	2)~Actions: The action set $ \cA $ is defined as the available aerial locations for the placement of the UAV-BS. At each time $ t $, the UAV-BS chooses an action $ a_t \in \cA $ for placement.
	3)~Rewards: The reward function is defined with the aim to maximize user throughput. Specifically, we assume that users are always handed over to the BS with the best SINR, so that an increased load at the UAV-BS usually indicates an increased user throughput due to better user SINRs. Hence, we define the reward to be the increased load at the UAV-BS. 
	
	We compare the proposed voting-based MARL algorithm with four baselines:
	1) the classic Q-learning algorithm~\cite{sutton2018reinforcement}, which uses centralized Q-learning to learn the optimal UAV placement policy;
	2) the multi-agent actor-critic algorithm based on the centralized learning with decentralized execution framework~\cite{lowe2017multi}, where distributed agents optimize the placement policy jointly by communicating with a centralized center;
	3) the multi-agent Q-learning algorithm proposed in~\cite{8807386}, where each agent performs independent Q-learning and treats the other agents as part of the environment;
	4) the optimal scheme, obtained by assuming that the underlying MDP is known.
	We refer to them as \textit{centralized QL}, \textit{multi-agent AC}, \textit{multi-agent QL}, and \textit{optimal} for short, respectively. 
	In addition, we adopt the majority voting rule proposed for multi-agent Q-learning in~\cite{Partalas2007Multi} to determine the joint action for both the multi-agent AC algorithm and the multi-agent QL algorithm.
	
	\begin{figure}[tb]
		\centering
		\includegraphics[trim = 1 1 1 1, clip, width=0.8\columnwidth]{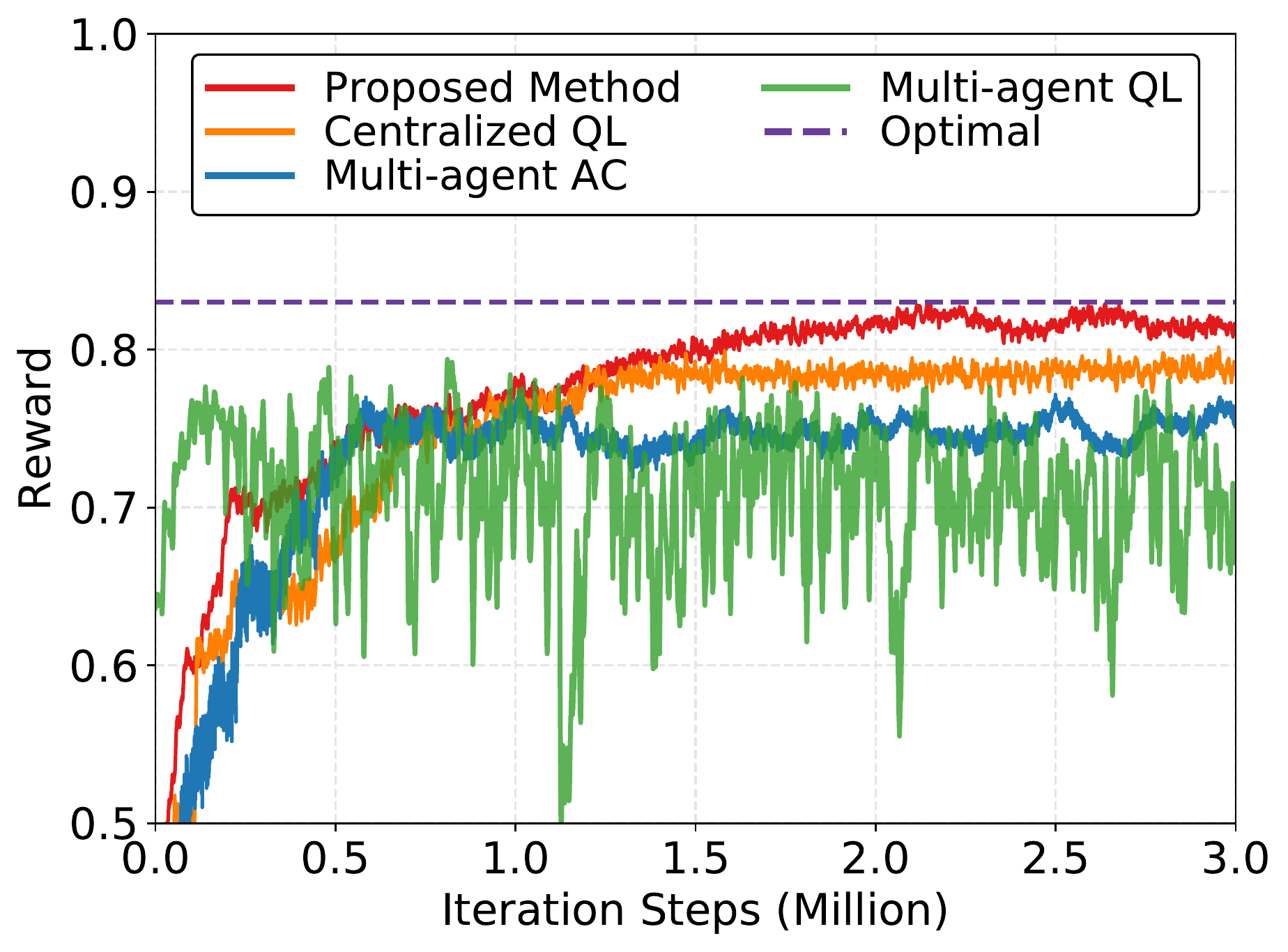}
		\caption{Rewards for the UAV-BS placement optimization.}
		\label{fig:experiment2}
	\end{figure}
	In Fig.~\ref{fig:experiment2}, we present the averaged rewards over $ 20 $ runs. The result shows that the performance of our proposed voting-based MARL algorithm outperforms all the comparing algorithms and is close to the optimal scheme.
	The discount factor for discounted RL methods is set to be $ 0.9 $. The performance gap between our proposed method and the centralized QL indicates that undiscounted RL methods are likely to outperform discounted RL methods in continuing optimization tasks. The performance gap between centralized QL and multi-agent AC/QL indicates that existing MARL algorithms exhibit a performance degradation compared with their centralized versions. In contrast, our proposed MARL algorithm achieves an equivalent performance to its centralized version. 
	In addition, the performance of the multi-agent QL algorithm is the worst and has a large variance. This verifies that specifying a proper collaboration protocol among the distributed agents is critical in MARL in order to improve the learning performance.

	\begin{figure*}[th]
		\centering
		\includegraphics[trim = 5 5 5 5, clip, width = 2\columnwidth]{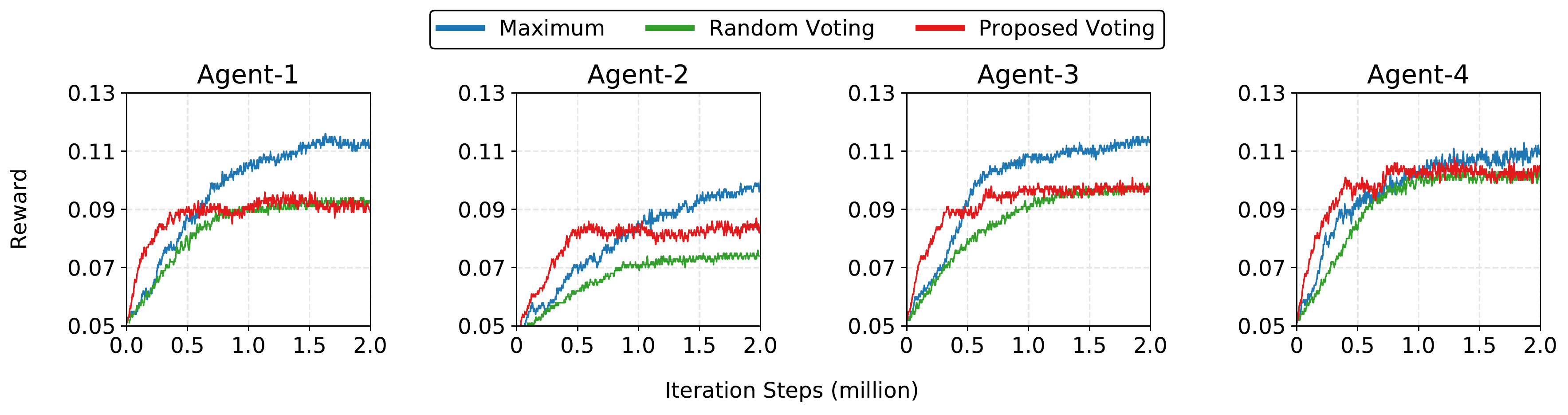}
		\caption{Local reward of each agent.}
		\label{fig:local_reward}
	\end{figure*}
	We further compare our proposed voting-based scheme with two baselines: 1) the random-voting scheme, where the MARL system randomly chooses one agent to determine the global action per iteration; 2) the greedy scheme, where the MARL system aims at maximizing the cumulative reward of a single agent.
	Fig.~\ref{fig:local_reward} presents the averaged reward of each agent over $ 20 $ runs. 
	The rewards of the greedy-maximizing scheme indicate the maximum obtainable reward of each agent, while the rewards of the random-voting scheme indicate the learning effectiveness without the proposed voting mechanism.
	The performance of our proposed voting-based scheme lies between the two baselines, which indicates that the agents are \textit{learning to compromise} in order to maximize the cumulative global reward. 
	%
	
	\section{Conclusions}
	\label{sec:conclusion}
	In this paper, we considered a collaborative MARL problem, where the agents vote to make group decisions. Specifically, the agents are coordinated to follow the proposed voting mechanism without revealing their own rewards to each other. 
	We gave a saddle-point formulation of the concerned MARL problem and proposed a primal-dual learning algorithm for solving it. We showed that our proposed algorithm achieves the same sublinear convergence rate as centralized learning. Finally, we provided empirical results to demonstrate the learning effectiveness.
	More interesting applications in the IoT system and the voting mechanism in the context of competitive MARL can be explored in the future. 
	
	\appendix[Proof of Theorem~\ref{thm:duality-gap}]
	Our proof shares a similar spirit as that of Theorem~1 in~\cite{wang2017primal}. However, the analysis in~\cite{wang2017primal} does not readily extend to the case of multi-agent AMDP. As a result, we have to develop a separate new convergence analysis here.
	
	By virtue of Lemma~\ref{lemma:equivalent-global}, it suffices to study the progress made by the sequences of global dual variables $\{\bm{\mu}^{g,t}\}_{t\ge0}$ and global primal variables $\{\bm{v}^t\}_{t\ge0}$ in Algorithm~\ref{algorithm:MARL}. We begin with the following lemma, which gives an estimate of the progress of the dual variables in terms of KL-divergence.
	\begin{Lemma}[Dual Improvement in KL-Divergence] \label{lemma-v}
		The iterates generated by Algorithm~\ref{algorithm:MARL} will satisfy
		\begin{equation} \label{eq:kl-improve}
		\begin{split}
		&\bbE\left[ D_{KL}({\bm{\mu}}^{g, *} \| \bm{\mu}^{g, t+1}) \mid \cF_t \right]  - D_{KL}(\bm{\mu}^{g, *} \| \bm{\mu}^{g, t}) \\
		&\leq \sumi \suma ({\mu}^{g,t}_{i,a} - \mu^{g,*}_{i,a} ) \bbE \left[   \Delta^{g,t}_{{i,a}} \mid \cF_t \right] \\
		&\quad+\frac{1}{2} \sumi \suma {\mu}^{g,t}_{{i,a}} \bbE \big[ \big( \Delta^{g,t}_{{i,a}} \big) ^2\mid \cF_t \big],
		\end{split}
		\end{equation}
		for all $t\geq 0$. \hfill$\blacksquare$
	\end{Lemma}
	{\smallskip\noindent\it Proof.} By definition, we have
	\begin{align*}
	& D_{KL}(\bm{\mu}^{g, *} \| \bm{\mu}^{g, t+1}) - D_{KL}(\bm{\mu}^{g, *} \| \bm{\mu}^{g, t}) \\
	&= \sumi\suma {\mu}^{g, *}_{{i,a}} \log \frac{{\mu}^{g, *}_{{i,a}}}{{\mu}^{g, t+1}_{{i,a}}} - \sumi\suma {\mu}^{g, *}_{{i,a}} \log \frac{{\mu}^{g, *}_{{i,a}}}{{\mu}^{g, t}_{{i,a}}} \\
	&= \sumi\suma {\mu}^{g, *}_{{i,a}} \log \frac{{\mu}^{g, t}_{{i,a}}} {{\mu}^{g, t+1}_{{i,a}}}.
	\end{align*}
	According to~\eqref{eq:local-xt},~\eqref{eq:global-dual-end}, and~\eqref{eq:glob-1}, we have
	\begin{align*}
	\log {\mu}^{g, t+1}_{{i,a}} &= \log \frac{\mu^{g, t}_{i,a} \exp \lbrace \Delta^{g, t}_{i,a} \rbrace}{\sumi \suma \mu^{g, t}_{i,a} \exp\lbrace \Delta^{g, t}_{i,a} \rbrace} \\
	&= \log \mu^{g, t}_{i,a} + \Delta^{g, t}_{i,a} - \log(Z),
	\end{align*}
	where $ Z =  \sumi \suma \mu^{g, t}_{i,a} \exp\lbrace \Delta^{g, t}_{i,a} \rbrace $. It follows that 
	\begin{align*}
	& D_{KL}(\bm{\mu}^{g, *} \| \bm{\mu}^{g, t+1}) - D_{KL}(\bm{\mu}^{g, *} \| \bm{\mu}^{g, t}) \\
	&= \sumi\suma {\mu}^{g, *}_{{i,a}} \log \frac{{\mu}^{g, t}_{{i,a}}} {{\mu}^{g, t+1}_{{i,a}}} \\
	&= \sumi\suma {\mu}^{g, *}_{{i,a}} \big( \log \mu^{g, t}_{i,a} - \log \mu^{g, t}_{i,a} - \Delta^{g, t}_{i,a} + \log(Z) \big) \\
	&= \log(Z)  - \sumi \suma {\mu}^{g, *}_{{i,a}} \Delta^{g, t}_{i,a}.
	\end{align*}
	Now, for any $\bm{v}^t \in \cV$, we have $\|\bm{v}^t\|_\infty \le 2t_{\text{mix}}^*$. Moreover, we have $r_{ij}^m(a) \in [0,1]$ by assumption. Hence, we have
	\[
	v_j^t - v_i^t + \sum_{m=1}^M r^m_{ij}(a) \le 4t^{*}_{\text{mix}} + M.
	\]
	This, together with the fact that $C= 4t^{*}_{\text{mix}} + M$, implies $ \Delta^{g,t}_{i,a} \leq 0 $, $\forall \ i \in \cS, \ a \in \cA, \ t=0,1,\ldots$. On the other hand,
	\begin{subequations}
		\begin{align}
		&\log(Z) \nonumber \\
		&= \log \left( \sumi \suma \mu^{g, t}_{i,a} \exp\big\lbrace \Delta^{g, t}_{i,a} \big\rbrace \right) \nonumber \\
		&\leq \log \sumi \suma {\mu}^{g,t}_{i,a} \left( 1+ \Delta^{g, t}_{{i,a}} + \frac12 \big(  \Delta^{g, t}_{{i,a}} \big) ^2 \right) \label{eq:exp-leq}\\
		&=\log \left( 1 +\ \sumi \suma {\mu}^{g,t}_{i,a}  \Delta^{g,t}_{i,a} + \frac12 \sumi \!\suma {\mu}^{g,t}_{i,a}  \big( \Delta^{g,t}_{i,a} \big)^{2} \right)  \nonumber\\
		&\leq \sumi \suma {\mu}^{g,t}_{i,a}  \Delta^{g,t}_{i,a} + \frac12 \sumi \suma {\mu}^{g,t}_{i,a}  \big( \Delta^{g,t}_{i,a} \big)^2, \label{eq:log-leq}
		\end{align}
	\end{subequations}
	where (\ref{eq:exp-leq}) uses the fact that $ \exp \left\lbrace x \right\rbrace \leq 1 + x + \frac12 x^2$ for $x \leq 0$ and (\ref{eq:log-leq}) uses the fact that $ \log(1+x) \leq x $ for $x>-1$. Therefore, by combining the above results and taking conditional expectation $ \bbE\left[ \cdot \mid \cF_t \right]  $ on both sides, we obtain~\eqref{eq:kl-improve}, as desired. \hfill$\blacksquare$
	
	\medskip
	Our strategy now is to bound the two terms on the right-hand side of~\eqref{eq:kl-improve} separately.
	
	\begin{Lemma} \label{lemma:lemma-KL}
		The iterates generated by Algorithm 1 satisfy
		\begin{align*}
		& \sumi \suma (\mu^{g,t}_{i, a} - \mu^{g,*}_{i, a}) \bbE \left[ \Delta^{g, t}_{i,a} \mid \cF_t \right] \\
		&= \frac{\beta}{|\cS|\cdot|\cA|}  \suma \left( \bm{\mu}^{g,t}_{a} - \bm{\mu}^{g,*}_{a}\right)^{\top} \left(  (P_a - I) \bv^t + \sum_{m=1}^M \bar{\boldsymbol{r}}^m_a \right) 
		\end{align*}
		for all $t\geq 0$. \hfill$\blacksquare$
	\end{Lemma}
	
	{\smallskip\noindent\it Proof.} 
	For arbitrary $i\in\cS$ and $a\in\cA$, we have
	\begin{align}
	&\sumi \suma (\mu^{g,t}_{i, a} - \mu^{g,*}_{i, a}) \bbE \left[ \Delta^{g, t}_{i,a} \mid \cF_t \right] \nonumber \\
	&= \frac{\beta}{|\cS|\cdot|\cA|} \sumi\suma ( \mu^{g,t}_{i, a} - \mu^{g,*}_{i, a} ) \left( (P_a - I) \bv^t + \sum_{m=1}^M \bar{\boldsymbol{r}}^m_a \right)_i  \nonumber \\
	&\quad - \frac{C\beta}{|\cS|\cdot|\cA|} \sumi\suma ( \mu^{g,t}_{i, a} - \mu^{g,*}_{i, a} ) \label{eq:step-1} \\
	&= \frac{\beta}{|\cS|\cdot|\cA|} \suma ( \bm{\mu}^{g,t}_{a} - \bm{\mu}^{g,*}_{a})^{\top} \left( ( P_a - I )\bv^t + \sum_{m=1}^M \bar{\boldsymbol{r}}^m_a \right), \label{eq:step-2}
	\end{align}
	where~\eqref{eq:step-1} follows from Lemma~\ref{lemma:unbiasedness} and~\eqref{eq:step-2} comes from the fact that 
	\[ 
	\sumi \suma \mu^{g,t}_{i,a} = \sumi \suma  \mu^{g,*}_{i,a} =1.
	\]
	This completes the proof. \hfill$\blacksquare$
	
	\medskip
	\begin{Lemma}\label{lemma-variance}
		The iterates generated by Algorithm 1 satisfy 
		\[ 
		\sumi \suma {\mu}^{g,t}_{i,a}  \bbE \big[ \big( \Delta^{g,t}_{{i,a}} \big) ^2\mid \cF_t \big] 
		\leq  \frac{4 \beta^2}{|\cS|\cdot|\cA|} \big( 4t^*_{\text{mix}} + M \big) ^2
		\] 
		for all $t\geq 0$. \hfill$\blacksquare$
	\end{Lemma}
	{\smallskip\noindent\it Proof.} Using~\eqref{key-0}, the assumptions that $r_{ij}^m(a) \in [0,1]$ and $\bm{v}^t \in \cV$, and the definition of $C$, we compute
	\begin{align*}
	& \bbE \big[ \big( \Delta^{g,t}_{{i,a}} \big) ^2\mid \cF_t \big] \\
	&=  \frac{\beta^2}{|\cS|\cdot|\cA|} \sumj p_{ij}(a) \left( v_j^t - v_i^t - C + \sum_{m=1}^M r^m_{ij}(a) \right) ^2 \\
	&\leq \frac{4\beta^2}{|\cS|\cdot|\cA|} \sumj p_{ij}(a)  \left( 4t^*_{\text{mix}} + M \right) ^2 \\
	&= \frac{4 \beta^2}{|\cS|\cdot|\cA|} \left( 4t^*_{\text{mix}} + M \right) ^2.
	\end{align*}
	Since $\sumi\suma \mu_{i,a}^{g,t+1}=1$, the result follows. \hfill$\blacksquare$
	
	\medskip
	Next, we give an estimate on the distance of the primal iterate $\bm{v}^t$ to the optimal primal variable $\bm{v}^*$.
	\begin{Lemma}[Distance to Primal Optimality] \label{lemma:lemma-v}
		The iterates generated by Algorithm~1 satisfy 
		\begin{align*}
		&\mathbb{E} \left[ \| \boldsymbol{v}^{t+1}-\boldsymbol{v}^*\|^2 \mid \mathcal{F}_t \right] \\
		&\leq \|\boldsymbol{v}^t - \boldsymbol{v}^*\|^2 + 2 {\alpha} (\boldsymbol{v}^t - \boldsymbol{v}^*)^{\top} 
		\left( \sum_{a \in \mathcal{A}}  (I-P_a)^{\top} \bm{\mu}_a^{g, t} \right) + 2\alpha^2
		\end{align*}
		for all $t\geq 0$. \hfill$\blacksquare$
	\end{Lemma}
	
	{\smallskip\noindent\it Proof.} 
	We compute
	\begin{align*}
	&\mathbb{E} \left[  \| \boldsymbol{v}^{t+1}-\boldsymbol{v}^*\|^2 \mid \mathcal{F}_t \right] \\
	&=  \mathbb{E} \left[ \| \Pi_{\mathcal{V}} \lbrace \boldsymbol{v}^t + \boldsymbol{d}^{t} \rbrace -\boldsymbol{v}^*\|^2 \mid \mathcal{F}_t \right] \\
	&\leq \mathbb{E} \left[ \| \boldsymbol{v}^t + \boldsymbol{d}^{t} - \boldsymbol{v}^* \|^2 \mid \mathcal{F}_t \right] \\
	&= \|\boldsymbol{v}^t - \boldsymbol{v}^*\|^2 + 2  (\boldsymbol{v}^t - \boldsymbol{v}^*)^{\top} \mathbb{E} \left[  \boldsymbol{d}^{t} \mid \mathcal{F}_t \right] + \mathbb{E} \left[  \|\boldsymbol{d}^{t} \|^2 \mid \mathcal{F}_t \right],
	\end{align*}
	where the inequality follows from the fact that $\bm{v}^* \in \cV$ and the projector $\Pi_{\cV}\{\cdot\}$ is non-expansive. By Lemma~\ref{lemma:unbiasedness}, we have
	\[ \mathbb{E} \left[ \bm{d}^{t} \mid  \mathcal{F}_t \right]  = \alpha \suma (I - P_a)^\top \bm{\mu}_a^{g,t}. \]
	Finally, using the definition of $\bm{d}^{t}$ in~\eqref{key-1}, we have $\mathbb{E} \left[ \| \boldsymbol{d}^{t} \|^2 \mid \mathcal{F}_t \right] = 2\alpha^2$. This completes the proof. \hfill$\blacksquare$
	
	\medskip
	We are now ready to establish the key recursion that will lead to our desired bound on the convergence rate of our proposed Algorithm~\ref{algorithm:MARL}.
	\begin{Lemma} \label{lemma:lemma-for-thm1}
		Define
		\begin{align*}
		V^t &= D_{KL}(\bm{\mu}^{g,*} \| \bm{\mu}^{g,t}) + \frac{1}{2 |\cS| (4\tmix+M)^2} \|\bv^t\!-\!\bv^*\|^2, \\
		W^{t} &= \suma (\bm{\mu}_a^{g,t})^{\top} \left( ( I - P_a ) \bv^* - \sum_{m=1}^M \bar{\boldsymbol{r}}^m_a \right) +  \bar v^*. 
		\end{align*}
		The iterates generated by Algorithm~1 satisfy 
		\[ 
		\bbE \left[ V^{t+1}\mid \cF_t \right] 
		\leq V^{t} -  \frac{\beta}{|\cS|\cdot|\cA|} W^{t} + 3\beta^2 \cdot \frac{(4\tmix + M)^2}{|\cS|\cdot|\cA|} 
		\] 
		for all $t\ge0$. \hfill$\blacksquare$
	\end{Lemma}
	
	{\smallskip\noindent\it Proof.} 
	Using the results in Lemmas~\ref{lemma-v}--\ref{lemma:lemma-v} and taking $ \alpha = \frac{1}{|\cA|} (4\tmix+M)^2 \beta $, we compute	
	\begin{align*}
	&\bbE \left[ V^{t+1} \mid \cF_t \right] \\
	&\leq V^{t} + \left( 2+\frac{1}{|\cA|} \right) \beta^2 \cdot \frac{(4\tmix+M)^2}{|\cS|\cdot|\cA|} \\
	&+ \frac{\beta}{|\cS|\cdot|\cA|} \suma \left( \bm{\mu}^{g,t}_{a} - \bm{\mu}^{g,*}_{a}\right)^{\top} \left(  (P_a - I) \bv^t + \sum_{m=1}^M \bar{\boldsymbol{r}}^m_a \right)  \\
	&+\frac{\beta}{|\cS|\cdot|\cA|} (\bv^t-\bv^*)^{\top} \left( \sum_{a \in \mathcal{A}}  (I-P_a)^{\top} \bm{\mu}_a^{g, t} \right).
	\end{align*}
	Now, observe that
	\begin{subequations}
		\begin{align}
		&\suma \left( \bm{\mu}^{g,t}_{a} - \bm{\mu}^{g,*}_{a}\right)^{\top} \left(  (P_a - I) \bv^t + \sum_{m=1}^M  \bar{\boldsymbol{r}}^m_a \right)  \nonumber \\
		&\quad+ (\bv^t-\bv^*)^{\top} \left( \sum_{a \in \mathcal{A}}  (I-P_a)^{\top} \bm{\mu}_a^{g, t} \right) \nonumber \\
		&=\suma \left( \bm{\mu}^{g,t}_{a} - \bm{\mu}^{g,*}_{a}\right)^{\top} \left(  (P_a - I) \bv^t + \sum_{m=1}^M  \bar{\boldsymbol{r}}^m_a \right)  \nonumber \\
		&\quad+ (\bv^t-\bv^*)^{\top} \left( \suma  (I - P_a)^{\top} ( \bm\mu_a^{g,t} - \bm\mu_a^{g,*}) \right) \label{eq:dual-feas} \\
		&=\suma \left( \bm{\mu}^{g,t}_{a} - \bm{\mu}^{g,*}_{a}\right)^{\top} \left( ( P_a - I) \bv^* + \sum_{m=1}^M \bar{\boldsymbol{r}}^m_a \right) \nonumber \\
		&=\suma (\bm{\mu}_a^{g,t})^{\top} \left( ( P_a - I ) \bv^* + \sum_{m=1}^M \bar{\boldsymbol{r}}^m_a \right) - \bar{v}^* \suma (\bm{\mu}_a^{g,*})^{\top} \boldsymbol{e} \label{eq:comple}
		\\ 
		&=\suma (\bm{\mu}_a^{g,t})^{\top} \left( ( P_a - I ) \bv^* + \sum_{m=1}^M \bar{\boldsymbol{r}}^m_a \right) - \bar v^*, \label{eq:sum-cond}
		\end{align}
	\end{subequations}
	where~\eqref{eq:dual-feas} and~\eqref{eq:sum-cond} use the dual feasibility conditions $\suma (\bm{\mu}^{g,*}_a)^\top (I-P_a) = \bm{0}$ and $\sumi\suma \mu^{g,*}_{i,a}=1$ in (\ref{eq:dual}), respectively;~\eqref{eq:comple} uses the complementarity condition 
	\[ 
	\mu_{i,a}^{g,*} \left( ( P_a - I ) \bv^* + \sum_{m=1}^M \bar{\boldsymbol{r}}^m_a - \bar v^*\cdot \boldsymbol{e} \right) _i = 0,\ \forall~i \in\cS, \ a\in\cA
	\]
	of the linear program (\ref{eq:primal}). Combining the preceding relations, we obtain Lemma \ref{lemma:lemma-for-thm1}. 
	\hfill$\blacksquare$
	
	{\smallskip\noindent\it Proof of Theorem~1:}
	We claim that 
	\[
	V^1 \leq \log(|\cS|\cdot|\cA|) + \frac{2(\tmix)^2}{(4\tmix+M)^2}.
	\]
	To see this, we note that ${\mu}^{g,1}$ is the uniform distribution and $\bv^0,\bv^*\in\cV$. Therefore, we have $D_{KL}(\bm{\mu}^{g,*} \| \bm{\mu}^{g,1}) \leq \log (|\cS|\cdot|\cA|)$ and $\|\bv^t- \bv^*\|^2\leq 4|\cS| (\tmix)^2$ for $t=0,1,\ldots$. This yields
	\begin{align*}
	V^1 &\leq  D_{KL}(\bm{\mu}^{g,*} \| \bm{\mu}^{g,1}) + \frac{1}{2|\cS| (4\tmix+M)^2 } \|\bv^1-\bv^*\|^2 \\
	&\leq \log(|\cS|\cdot|\cA|) + \frac{2(\tmix)^2}{(4\tmix+M)^2}.
	\end{align*}
	Now, we rearrange the terms in Lemma \ref{lemma:lemma-for-thm1} and obtain
	\[ 
	W^t \leq \frac{|\cS|\cdot|\cA|}{\beta}(V^t - \bbE \left[  V^{t+1} \mid \cF_t \right]  ) + 3\beta  (4\tmix+M)^2.
	\] 
	Summing over $t=1,\ldots,T$ and taking the expectation, we have 
	\begin{align*}
	&\bbE \left[  \sum_{t=1}^{T} W^t \right] \\
	& \leq \frac{|\cS|\cdot|\cA|}{ \beta  }\sum_{t=1}^{T}( \bbE \left[ V^t \right] - \bbE \left[ V^{t+1} \right] )+ 3\beta T (4\tmix+M)^2  \\
	&= \frac{ |\cS|\cdot|\cA|}{\beta } \left( \bbE \left[ V^1 \right] - \bbE \left[ V^{T+1} \right] \right)  + 3\beta T (4\tmix+M)^2  \\
	&\leq \frac{|\cS|\cdot|\cA|}{\beta }  \left( \log(|\cS|\cdot|\cA|) + \frac{2(\tmix)^2}{(4\tmix+M)^2} \right) \\
	&\quad+ 3\beta T (4\tmix+M)^2.
	\end{align*}
	By taking 
	\[ \beta  =\frac{1}{4\tmix+M} \sqrt{ \frac{|\cS|\cdot|\cA| \cdot  \log(|\cS|\cdot|\cA|) }{2 T} }, \]
	we obtain
	\[
	\bbE \left[ \frac{1}{T}\sum_{t=1}^{T} W^t \right] = \tilde{O} \left( \left( 4\tmix+M \right)  \sqrt{\frac{ |\cS|\cdot|\cA| }{T}}\right),
	\]
	as desired. \hfill$\blacksquare$
	
	\bibliographystyle{IEEEtran}
	\bibliography{../bib/IEEEabrv2015,../bib/MARL_Journal}
\end{document}